# Enhanced Optimization with Composite Objectives and Novelty Selection

Hormoz Shahrzad[1], Daniel Fink[1] and Risto Miikkulainen[1,2]

[1] Sentient Technologies
[2] University of Texas at Austin
hormoz@sentient.ai

**Abstract**

An important benefit of multi-objective search is that it maintains a diverse population of candidates, which helps in deceptive problems in particular. Not all diversity is useful, however: candidates that optimize only one objective while ignoring others are rarely helpful. This paper proposes a solution: The original objectives are replaced by their linear combinations, thus focusing the search on the most useful tradeoffs between objectives. To compensate for the loss of diversity, this transformation is accompanied by a selection mechanism that favors novelty. In the highly deceptive problem of discovering minimal sorting networks, this approach finds better solutions, and finds them faster and more consistently than standard methods. It is therefore a promising approach to solving deceptive problems through multi-objective optimization.

## 1-Introduction

Multi-objective optimization is most commonly useful in discovering a Pareto front from which solutions that represent useful tradeoffs between objectives can be selected (Coello Coello, 2007; Deb et al. 2002; Deb and Jain, 2014; Deb et al. 2016; Jain and Deb, 2014). Evolutionary methods are a natural fit for such problems because the Pareto front naturally emerges in the population maintained in these methods. Interestingly, multi-objectivity can also improve evolutionary optimization because it encourages populations with more diversity. Even when the focus of optimization is find good solutions along a primary performance metric, it is useful to create secondary dimensions that reward solutions that are different, e.g. in terms of structure, size, cost, consistency etc. Multi-objective optimization then discovers stepping stones that can be combined to achieve high fitness along the primary dimension (Meyerson and Miikkulainen, 2017). The stepping stones are useful in particular in problems where the fitness landscape is deceptive, i.e. where the optima are surrounded by inferior solutions (Lehman and Miikkulainen, 2014).

However, not all such diversity is useful. In particular, candidates that optimize one objective only and ignore the others are less likely to lead to useful tradeoffs, and are less likely to escape deception. The main idea evaluated in this paper is to replace the objectives with their linear combinations, thus focusing the search in more useful areas of the search space. In effect, the Pareto axes become angled, and search focuses more on tradeoffs instead of single objectives, allowing it to search around deceptive areas.

Naturally, some diversity is lost with such a focus. The second idea in this paper is that diversity can be encouraged more directly in the remaining space by utilizing a novelty metric for selection. Among the best candidates, those that are most different from the others are selected for reproduction; among the worst candidates, those that are the least different from the others will be discarded. Such a bias for diversity creates synergetic focus on tradeoffs. Together they result in a powerful method for optimization in domains where a primary performance objective can be supplemented with secondary objectives for diversity.

These ideas are tested in this paper in the highly deceptive domain of sorting networks (Knuth, 1998), i.e. networks of comparators that map any set of numbers represented in their input lines to a sorted order in their output lines. These networks have to be correct, i.e. sort all possible cases of input. The goal is to discover networks that are as small as possible, i.e. have as few comparators organized in as few sequential layers as possible. While correctness is the primary objective, it is actually not that difficult to achieve, because it is not deceptive. Minimality on the other hand, is highly deceptive and makes the sorting network design an interesting benchmark problem.

The composite novelty method is implemented in this domain and evaluated in four steps. As a baseline, a single objective combining correctness and minimality is first run. It lacks diversity and is effective only with the simplest networks. Second, the standard multi-objective approach is then implemented based with NSGA-II (Deb et al. 2002), with inaccuracy, number of layers, and number of comparators as the dimensions to be minimized. The approach has increased diversity, and finds solutions faster and to harder problems, but it also finds many solutions that are not useful. Third, these objectives are replaced with composites: one objective consists primarily of inaccuracy, with some layer and comparator fitness included; the other two consist a proportional combination of primarily layer and comparator fitness, with some correctness included. The solutions are found even faster and more consistently, but they are not yet optimal quality, presumably due to lost diversity in search. Fourth, novelty-based selection is included in the method, improving the search and resulting in solutions with better quality. This method finds optimal or near-optimal solutions to sorting networks with 8 to16 lines, and could likely find more with more extensive computational resources.

The composite novelty method is thus a promising approach to a range of problems where secondary objective is available to diversify search.

# 2-Background and Related Work

Evolutionary methods for optimizing single-objective and multi-objective problems are discussed, as well as the idea of using novelty to encourage diversity. The problem of minimal sorting networks is introduced and prior work in it reviewed.

## 2.1 Single-objective optimization

When the optimization problem has a smooth and non-deceptive search space, evolutionary optimization of a single objective is usually convenient and effective. However, we are increasingly faced with problems with more than one objective and with a rugged and deceptive search space. The first approach often is to combine the objectives to a composite version:

$$\text{Composite}(O_1, O_2, \ldots, O_k) = \sum_{i=1}^{k} \alpha_i O_i^{\beta_i}, \quad (1)$$

Where the constant hyper-parameters $\alpha_i$ and $\beta_i$ determine the relative importance of each objective in the composition.

The parameterization of a composite objective can be done in two ways:
1. By folding the objective space, and thereby causing multitude of solutions to have the same value. Diversity is lost since solutions with different behavior are considered to be equal.
2. By creating a hierarchy in the objective space, and thereby causing some objectives to have more impact than many of the other objectives combined. The search will thus optimize the most important objectives first, which in deceptive domains might not be the best way, or possible at all. These problems can be avoided by casting the problem explicitly as multi-objective optimization.

## 2.2 Multi-objective optimization

In contrast, multi-objective optimization methods construct a Pareto set of solutions (Deb et al. 2016), and therefore eliminate the issues with objective folding and hierarchy. However, not all diversity in the Pareto space is useful. Candidates that optimize one objective only and ignore the others are less likely to lead to useful tradeoffs, and are less likely to escape deception.

One potential solution is reference-point based multi-objective methods such as NSGA-III (Deb et al. 2016; Deb and Jain, 2014). They make it possible to harvest the tradeoffs between many objectives and can therefore be used to select for useful diversity as well, although they are not as clearly suited for escaping deception.

An alternative, proposed in this paper, is to use composite multi-objective axes to focus the search on the area with most useful tradeoffs. Since the axes are not orthogonal, solutions that optimize only one objective will not be on the Pareto front. The focus effect, i.e. the angle between the objectives, can be tuned by varying the coefficients of the composite.

However, focusing the search in this manner has the inevitable side effect of reducing diversity. Therefore, it is important that the search method makes use of whatever diversity exists in the focused space. Incorporating a preference for novelty does exactly that.

## 2.3 Novelty search

Novelty search (Lehman and Stanley, 2011; Lehman and Stanley, 2008) is an increasingly popular paradigm that overcomes deception by ranking solutions based on how different they are from others. Novelty is computed in the space of behaviors, i.e., vectors containing semantic information about how a solution achieves its performance when it is evaluated. However, with a large space of possible behaviors, novelty search can become increasingly unfocused, spending most of its resources in regions that will never lead to promising solutions.

Recently, several approaches have been proposed to combine novelty with a more traditional fitness objective (Gomes et al. 2015; Gomes, 2009; Mouret, 2011; Mouret and Doncieux, 2012; Pugh et al. 2015) to reorient search towards fitness as it explores the behavior space. These approaches have helped scale novelty search to more complex environments, including an array of control (Bowren et al. 2016; Cully et al. 2015; Mouret and Doncieux, 2012) and content generation (Lehman et al. 2016; Lehman and Stanley, 2012; Lehman and Stanley, 2011) domains.

Many of these approaches combine a fitness objective with a novelty objective in some way, for instance as a weighted sum (Cuccu and Gomez, 2011), or as different objectives in a multi-objective search (Mouret, 2011; Mouret and Doncieux, 2012). Another approach is to keep the two kinds of search separate, and make them interact through time. For instance, it is possible to first create a diverse pool of solutions using novelty search, presumably overcoming deception that way, and then find solutions through fitness-based search (Krcah, and Toropila, 2010). A third approach is to run fitness-based search with a large number of objective functions that span the space of solutions, and use novelty search to encourage search to utilize all those functions (Cully et al. 2015; Mouret and Clune. 2015; Pugh et al. 2015). A fourth category of approaches is to run novelty search as the primary mechanism, and use fitness to select among the solutions. For instance, it is possible to add local competition through fitness to novelty search (Lehman and Stanley, 2011). Another version is to accept novel solutions only if they satisfy minimal performance criteria (Gomes et al. 2015; Lehman and Stanley, 2010). Some of these approaches have been generalized using the idea of behavior domination to discover stepping stones (Meyerson and Miikkulainen, 2017; Meyerson et al. 2016).

This paper takes a slightly different approach. Since multiple objectives are used as the primary driver of novelty, and the goal is to make sure the multi-objective space is searched thoroughly, novelty is used simply in selecting which individuals to reproduce and which to discard. This combination is particularly effective, as the experiments in the sorting network domain will demonstrate.

## 2.4 Sorting networks

A sorting network of n inputs is a fixed layout of comparison-exchange operations (comparators) that sorts all inputs of size n (Figure 1) (Knuth, 1998). Since the same layout can sort any input, it represents an oblivious or data-independent sorting algorithm, that is, the layout of comparisons does not depend on the input data. The resulting fixed communication pattern makes sorting networks desirable in parallel implementations of sorting, such as those in graphics processing units, multi-processor computers, and switching networks (Baddar, 2009; Kipfer et al. 2004; Valsalam and Miikkulainen, 2013). Beyond validity, the main goal in designing sorting networks is to minimize the number of layers, because it determines how many steps are required in a parallel implementation. A tertiary goal is to minimize the total number of comparators in the networks. Designing such minimal sorting networks is a challenging optimization problem that has been the subject of active research since the 1950s (Knuth, 1998). Although the space of possible networks is infinite, it is relatively easy to test whether a particular network is correct: If it sorts all combinations of zeros and ones correctly, it will sort all inputs correctly (Knuth, 1998).

Many of the recent advances in sorting network design are due to evolutionary methods (Valsalam and Miikkulainen, 2013). However, it is still a challenging problem even for the most powerful evolutionary methods because it is highly deceptive: Improving upon a current design may require temporarily growing the network, or sorting fewer inputs correctly. Sorting networks are therefore a good domain for testing the power of evolutionary algorithms.

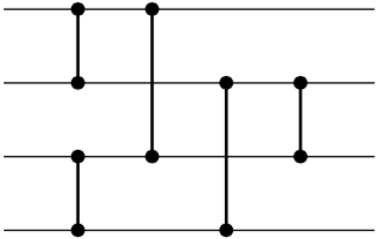

**Figure 1**: A Four-Input Sorting Network. This network takes as its input (left) four numbers, and produces output (right) where those number are sorted (large to small, top to bottom). Each comparator (connection between the lines) swaps the numbers on its two lines if they are not in order, otherwise it does nothing. This network has three layers and five comparators, and is the minimal four-input sorting network. Minimal networks are generally not known for large input sizes and designing them is a challenging optimization problem.

## 3-Methods

The representation of the sorting network domain and the comparison setup is first described, followed by the single and multi-objective optimization methods, the composite objective method, and novelty-based selection method.

### 3.1 Representing sorting networks

Because this paper focuses on evaluating the composite novelty method, a general representation of the sorting network problem, to which various evolutionary techniques can be readily applied, was developed. In this representation, sorting networks of $n$ lines are seen as a sequence of two-leg comparators where each leg is connected to a different input line and the first leg is connected to a higher line than the second:

$$\{(f_1, s_1), (f_2, s_2), (f_3, s_3), \dots, (f_c, s_c)\}.$$

The number of layers can be determined from such a sequence by grouping successive comparators together into a layer until the next comparator would add a second connection to one of the lines in the same layer. With this representation, mutation and crossover operators amount to adding and removing a comparator, swapping two comparators, and crossing over the comparator sequences of two parents at a single point.

Domain-specific techniques such as mathematically designing the prefix layers (Codish et al. 2014 and 2016) or utilizing certain symmetries (Valsalam and Miikkulainen, 2013) were not used (they can be used in the future to improve the results further). The experiments were also standardized to a single machine (a multi-core desktop) with no cloud or distributed evolution benefits (Hodjat et al. 2016). To facilitate comparisons, a pool of one thousand individuals were evolved for thousand generations with each method.

### 3.2 Single-objective approach

In order to design an effective objective for the single-objective approach, note that correctness is part of the definition of a sorting network: Even if a network mishandles only one sample, it will not be useful. The number of layers can be considered the most important size objective because it determines the efficiency of a parallel implementation. A hierarchical composite objective can therefore be defined as:

$$\text{SingleFitness}(m, l, c) = 10000\, m + 100\, l + c, \qquad (2)$$

Where $m, l,$ and $c$ are the number of mistakes (unsorted samples), number of layers, and number of comparators, respectively.

In the experiments in this paper, the solutions will be limited to less than hundred layers and comparators, and therefore, the fitness will be completely hierarchical (i.e. there is no folding).

### 3.3 Multi-objective approach

In the multi-objective approach the same dimensions, i.e. the number of mistakes, layers, and comparators $m, l, c$, are used as three separate objectives. They are optimized by the NSGA-II algorithm (Deb et al. 2002) with selection percentage of 10%. Indeed this approach may discover solutions with just a single layer, or a single comparator, since they qualify for the Pareto front. Therefore, diversity is increased compared to the single-objective method, but not necessarily helpful diversity.

### 3.4 Composite multi-objective approach

In order to construct composite axes, each objective is augmented with sensitivity to the other objectives:

$$\text{Composite}_1(m, l, c) = 10000\, m + 100\, l + c, \qquad (3)$$

$$\text{Composite}_2(m, l) = \alpha_1 m + \alpha_2 l, \qquad (4)$$

$$\text{Composite}_3(m, c) = \alpha_3 m + \alpha_4 c. \qquad (5)$$

The primary composite objective (Formula 3), which will replace the mistake axis, is the same hierarchical fitness used in the single-objective approach. It discourages evolution from constructing correct networks that are extremely large. The second objective (Formula 4), with $\alpha_2 = 10$, primarily encourages evolution to look for solutions with a small number of layers. A much smaller cost of mistakes, with $\alpha_1 = 1$, helps prevent useless single-layer networks from appearing in the population. Similarly, the third objective

(Formula 5), with $\alpha_3 = 1$ and $\alpha_4 = 10$, applies the same principle to the number of comparators.

These values for $\alpha_1, \alpha_2, \alpha_3,$ and $\alpha_4$ were found to work well in this application, but the approach is not very sensitive to them; A broad range will work as long as they establish a primacy relationship between the objectives. Also, even though the composite multi-objective approach introduces these additional hyper parameters, they do not usually require significant tuning. Their values arise naturally from the problem domain based on how some solutions are preferred over others. For example, in the sorting network domain the values can easily be set to push system toward prioritizing number of layers over comparators if so desired.

### 3.5 Novelty selection method

In order to measure how novel the solutions are it is first necessary to be able to characterize their behavior. While there are many ways to do it, a concise and computationally efficient way is to count how many swaps took place on each line in sorting all possible zero-one combinations during the validity check. Such a characterization is a vector that has the same size as the problem, making the distance calculations very fast. It also represents the true behavior of the network; that is, even if two networks sort the same input cases correctly, they may do it in different ways, and the characterization is likely to capture that difference. Given this behavior characterization, novelty of a solution is then measured by the sum of pairwise distances of its behavior vector to those of all the other individuals in the selection pool:

$$\text{NoveltyScore}(x_i) = \sum_{j=1}^{n} d(b(x_i), b(x_j)). \quad (6)$$

The selection method also has another parameter called selection multiplier (e.g. two in these experiments), varying between one and the inverse of the elite fraction (e.g. 1/10, i.e. 10%) used in the NSGA-II multi-objective optimization method. The original selection percentage is multiplied by the selection multiplier to form a broader selection pool. That pool is sorted according to novelty, and the top fraction representing the original selection percentage is used for selection. This way, good solutions that are more novel are included in the pool.

One potential issue is that a cluster of solutions far from the rest may end up having high novelty scores while only one of them is good enough to keep. Therefore, after the top fraction is selected, the rest of the sorted solutions are added to the selection pool one by one, replacing the solution with the lowest minimum novelty, defined as

$$\text{MinimumNovelty}(x_i) = \min_{1 \leq j \leq n; \, j \neq i} d(b(x_i), b(x_j)). \quad (7)$$

Note that this method allows tuning novelty selection continuously between two extremes: by setting it to one, the method reduces to the original multi-objective method (i.e. only the elite fraction ends up in the final elitist pool), and by setting it to the inverse of the elite fraction reduces it to pure novelty search (i.e. the whole population, sorted by novelty, is the selection pool) In practice, low and midrange values work well, including the value two used in these experiments.

## 4-Experiments

The methods were evaluated in the problem of discovering minimal sorting networks, and results evaluated in terms of correctness and minimization.

### 4.1 Experimental setup

In order to evaluate the composite novelty method, 480 experiments were run with the following parameters:
- Four methods tested (Single Objective, Multi-Objective, Composite Multi-Objective, and Composite Multi-Objective Novelty; Multi-Objective Novelty was excluded because it showed no comparable improvements in preliminary experiments).
- Twelve network sizes (5 through 16)
- Ten repetitions for each configuration
- Population of one thousand for the pool
- A thousand generations runtime
- 10% elitist selection

Method-specific parameters were specified above in subsections of section 3.

### 4.2 Correctness

All 480 experiments were able to find solutions that sort all inputs correctly. Indeed, it is relatively easy to keep adding comparators until the network sorts everything correctly; there is little deception. The challenge comes from having to do it with minimal comparators and layers: Removing a comparator may require changing the network drastically to make it still sort correctly. Thus, although minimization is a secondary goal in constructing sorting networks, it is actually the more challenging one.

### 4.3 Minimization

Minimization performance of the four methods is illustrated in Figures 2-5; the smallest known solution is also plotted for comparison (lower is better).

The five-line sorting problem is simple enough so that all methods were able to discover optimal solutions in all runs. The methods' performance started to diverge from six lines on, and the differences became more pronounced the larger the problem.

Figure 2 shows the best runs in terms of comparators, and Figure 3 in terms of number of layers. The Composite Multi-Objective Novelty method performs the best, followed by Composite Multi-Objective, Multi-Objective, and Single-Objective method.

The average results follow a similar pattern. Figure 4 shows the number of comparators and Figure 5 the average number of layers in the best solutions found, averaged over the ten runs. Again, the Composite Multi-Objective Novelty method performs the best, followed by Composite Multi-Objective, Multi-Objective, and Single-Objective methods. In terms of statistical significance ($p<0.05$), the Multi-Objective approach achieves significant improvement over Single-Objective at 16-lines networks, while Composite Multi-Objective significantly outperforms Multi-Objective all the way from 9-lines to 16 lines. Composite Multi-Objective Novelty is better than Composite Multi-Objective in most networks after 11-lines.

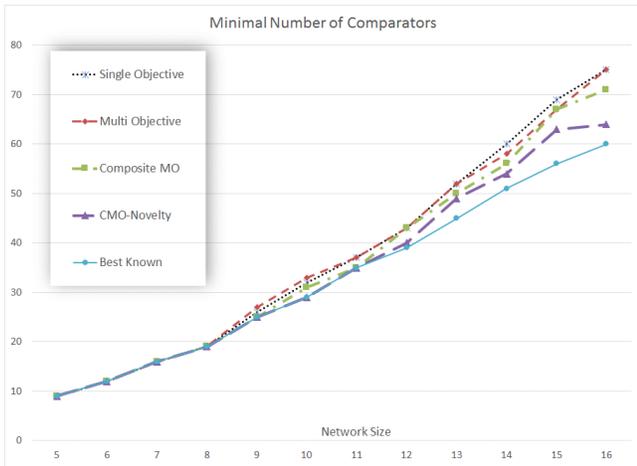

Figure 2: The minimal number of comparators discovered in the best run of each method over different size problems.

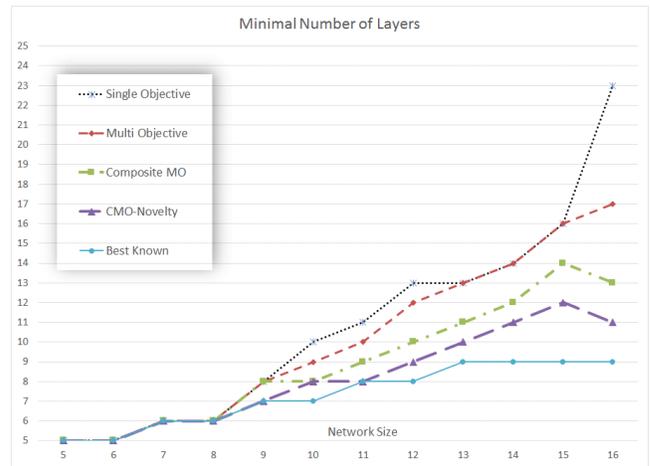

Figure 3: The minimal number of layers discovered in the best run of each method over different size problems.

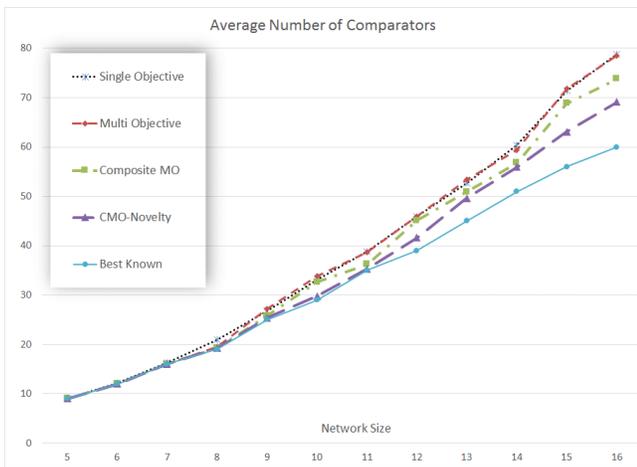

Figure 4: The average minimal number of comparators discovered by each method over ten runs.

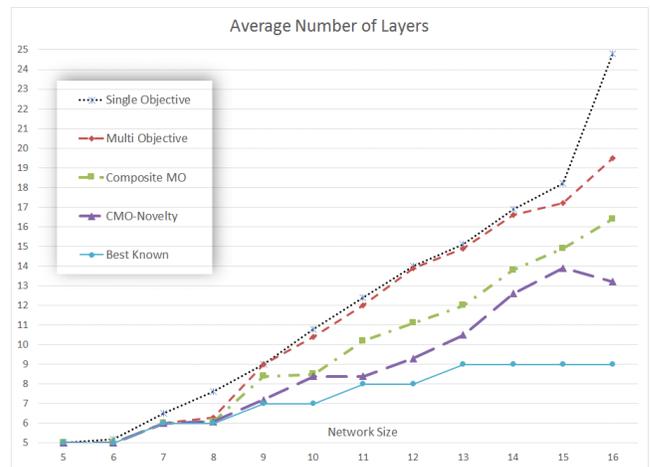

Figure 5: The average minimal number of layers discovered by each method over ten runs.

The results thus validate the ideas behind these methods: Each innovation is an improvement over the preceding one. The plots also show that there is still room for improvement. Indeed, the runs were limited to thousand generations to facilitate comparisons in this paper. In many cases, the results were still improving, and it is indeed the nature of this problem that longer runs give better results, as long as diversity can be maintained. Such experiments constitute a compelling direction for future work.

## Discussion and Future Work

The results in the minimal sorting network domain illustrate the principles employed in composite novelty approach well. The secondary objectives diversify the search, composite objectives focus it on most useful areas, and novelty selection establishes a thorough exploration in those areas. These methods are readily implemented in standard multi-objective search such as NSGA-II, and can be used in combination of many other techniques already developed to improve evolutionary multi-objective optimization.

The sorting network experiments were designed to demonstrate the potential of the method, but they do not yet illustrate its full power. One compelling direction of future work is to use it to optimize sorting networks systematically, with domain-specific techniques integrated into the search, and with significantly more computing power. It is likely that given such power, many new minimal networks can be discovered. (At the time of this writing, longer runs of CMO-Novelty have matched all known best results up to 18 lines.)

The method can also be applied in many other domains, in particular those that are deceptive and have natural secondary objectives. For instance various game strategies from board to video games can be cast in this form, where winning is accompanied by different dimensions of the score. Solutions for many design problems, such as 3D printed objects, need to satisfy a set of functional requirements, but also maximize strength and minimize material. Effective control of robotic systems need to accomplish a goal while minimize energy and wear and tear. Thus, many applications should be amenable to this approach.

Another direction is to extend the method further into discovering effective collections of solutions. For instance, ensembling is a good approach for increasing the performance of machine learning systems. Usually the ensemble is formed from solutions with different initialization or training, with no mechanism to ensure that their differences are useful. In composite novelty, the Pareto front consists of a diverse set of solutions that span the area of useful tradeoffs. Such collections should make for a powerful ensemble, extending the applicability of the approach.

# Conclusion

The composite novelty method is a promising approach to deceptive problems where a secondary objective is available to diversify the search. In such cases, composite objectives focus the search on the most useful tradeoffs and allow escaping deceptive areas. Novelty-based selection increases exploration in the focus area, leading to better solutions, faster and more consistently and it can be combined with almost any fitness based method. These principles were demonstrated in this paper in the highly deceptive problem of minimizing sorting networks, but they should apply to many other problems of the same kind, thus increasing the power and reach of evolutionary multi-objective optimization.